\documentclass[sigconf]{acmart}

\AtBeginDocument{%
  }

\copyrightyear{2024} 
\acmYear{2024} 
\setcopyright{acmlicensed}\acmConference[ICAIF '24]{5th ACM International Conference on AI in Finance}{November 14--17, 2024}{Brooklyn, NY, USA}
\acmBooktitle{5th ACM International Conference on AI in Finance (ICAIF '24), November 14--17, 2024, Brooklyn, NY, USA}
\acmDOI{10.1145/3677052.3698597}
\acmISBN{979-8-4007-1081-0/24/11}

\usepackage{hyperref}
\usepackage{url}
\usepackage{tikz}
\usepackage{multirow}
\usepackage{xspace}
\usepackage{booktabs}

\usepackage{graphicx} 
\usepackage{caption}
\usepackage{enumitem}
\usepackage{verbatim}
\usepackage{listings}
\usepackage{xcolor}
\usepackage{scalerel} 
\usepackage{float}
\usepackage{subcaption}
\usepackage[inkscapeformat=png]{svg}
\usepackage{bbding}
\usepackage{comment}
\usepackage{tabularray}
\usepackage{booktabs,tabularx}
\usepackage{stfloats}

\begin{document}

\title[FISHNET: Financial Intelligence from Swarms]{FISHNET: Financial Intelligence from Sub-querying, Harmonizing, Neural-Conditioning, Expert Swarms, and Task Planning}

\author{Nicole Cho}
\affiliation{%
  \institution{J.~P.~Morgan AI Research}
  \city{New York}
  \state{NY}
  \country{USA}
}
\email{nicole.cho@jpmorgan.com}
\orcid{0009-0006-9007-3255}

\author{Nishan Srishankar}
\affiliation{%
  \institution{J.~P.~Morgan AI Research}
  \city{New York}
  \state{NY}
  \country{USA}
}
\email{nishan.srishankar@jpmchase.com}
\orcid{0000-0002-7385-3018}

\author{Lucas Cecchi}
\affiliation{%
  \institution{J.~P.~Morgan AI Research}
  \city{New York}
  \state{NY}
  \country{USA}
}
\email{lucas.cecchi@jpmchase.com}
\orcid{0009-0003-4501-3710}

\author{William Watson}
\affiliation{%
  \institution{J.~P.~Morgan AI Research}
  \city{New York}
  \state{NY}
  \country{USA}
}
\email{william.watson@jpmchase.com}
\orcid{0000-0001-5516-262X}

\renewcommand{\shortauthors}{Cho et al.}

\begin{abstract}
Financial intelligence generation from vast data sources has typically relied on traditional methods of knowledge-graph construction or database engineering. Recently, fine-tuned financial domain-specific Large Language Models (LLMs), have emerged. While these advancements are promising, limitations such as high inference costs, hallucinations, and the complexity of concurrently analyzing high-dimensional financial data, emerge. This motivates our invention FISHNET (Financial Intelligence from Sub-querying, Harmonizing, Neural-Conditioning, Expert swarming, and Task planning), an agentic architecture that accomplishes highly complex analytical tasks for more than 98,000 regulatory filings that vary immensely in terms of semantics, data hierarchy, or format. FISHNET shows remarkable performance for financial insight generation (61.8\% success rate over 5.0\% Routing, 45.6\% RAG R-Precision). We conduct rigorous ablations to empirically prove the success of FISHNET, each agent's importance, and the optimized performance of assembling all agents. Our modular architecture can be leveraged for a myriad of use-cases, enabling scalability, flexibility, and data integrity that are critical for financial tasks.
\end{abstract}

\ccsdesc[500]{Applied computing~Economics}
\ccsdesc[500]{Information systems~Information retrieval query processing}
\ccsdesc[500]{Information systems~Information extraction}

\keywords{LLM Agents, Swarming, Harmonizing, Planning, Sub-querying}


\maketitle

\begin{figure}[t!]
  \centering
  \includegraphics[clip, trim=5cm 0.1cm 5.3cm 0.1cm, width=0.52\textwidth]{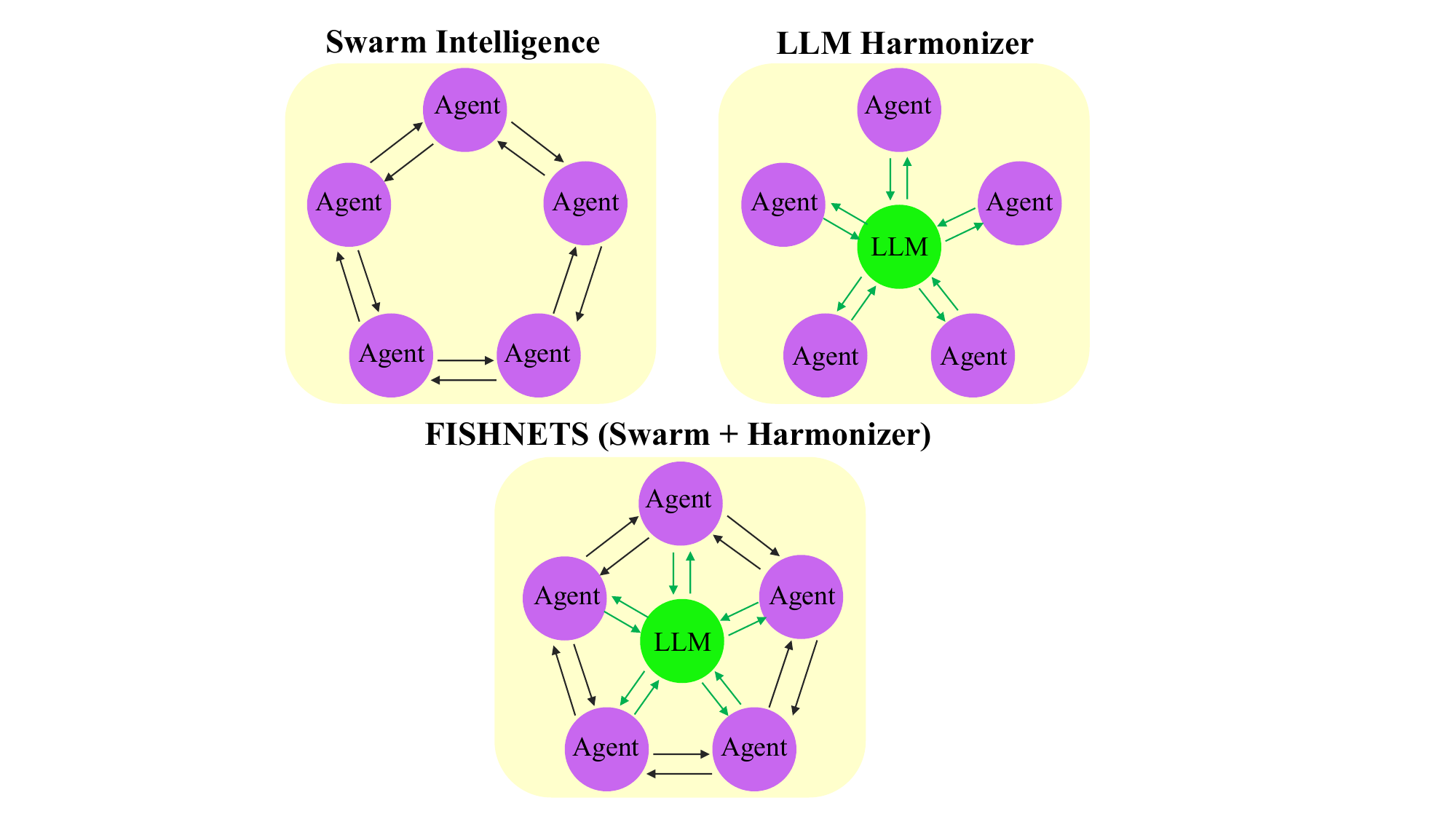}
  \caption{Traditional Swarm Intelligence (SI) vs. FISHNET: SI relies on highly capable individual agents collectively working towards an efficient solution that exceeds the respective capabilities of those agents. SI typically excludes the presence of a central entity that harmonizes the agents' actions. Recently, a plethora of studies have explored an LLM's capability to orchestrate actions. In FISHNET, we combine these two approaches together; a central harmonizer can orchestrate while the expert agents also communicate. 
  }
  \label{fig:teaser}
  \Description{Teaser Image for the comparison of prior work with our proposed system: FISHNET.}
\end{figure}

\section{Introduction}

\begin{figure*}[t!]
  \centering
  \includegraphics[clip, trim=1.15cm 0.12cm 1.2cm 0.1cm, width=1.0\textwidth]{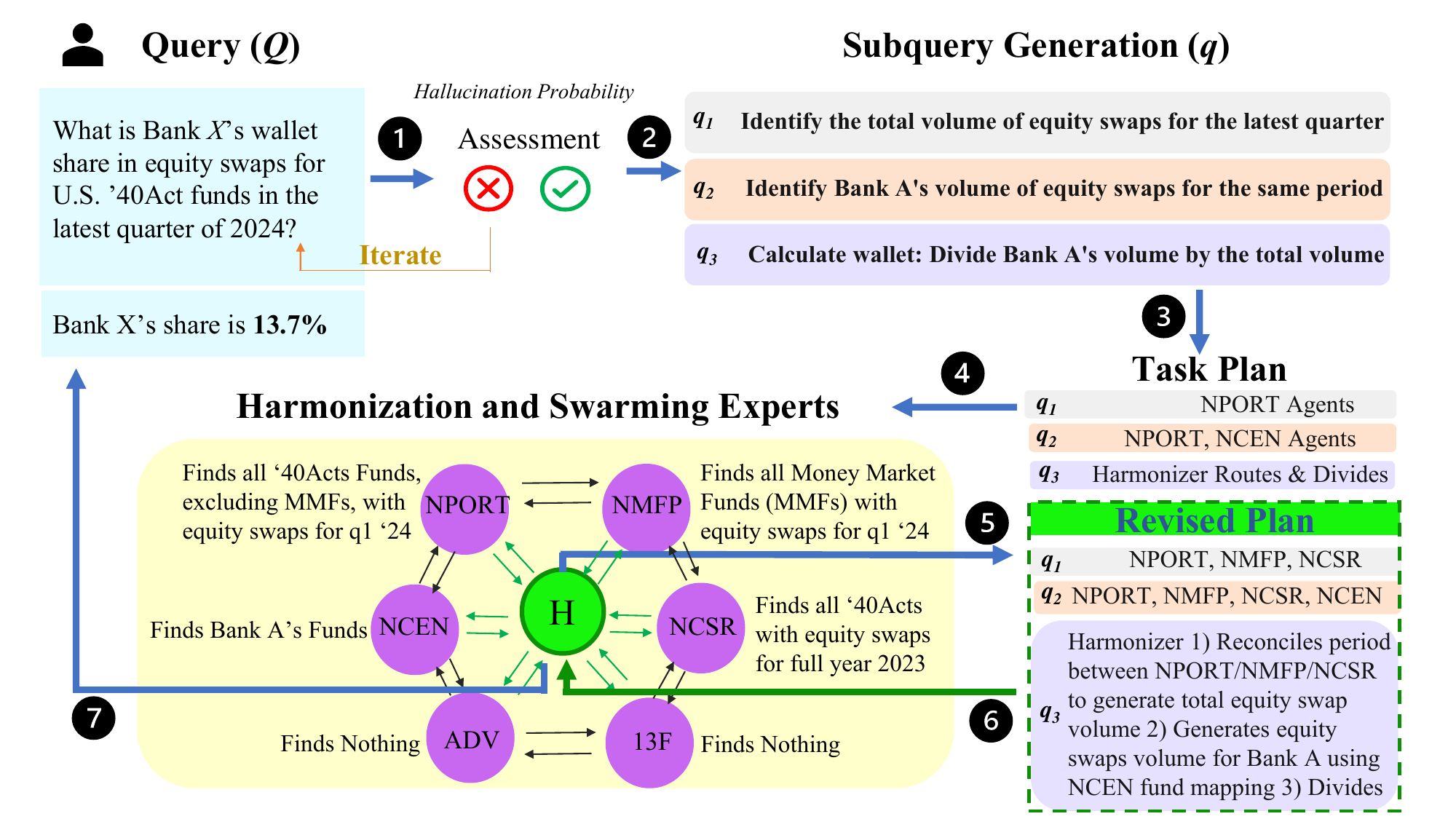}
  \caption{System Diagram of FISHNET: For any imposed query, (1) the query's hallucination probability is firstly calculated to improve on the query's quality. (2) Then, our sub-querying agent breaks down the query into different components. (3) Our task planning agent, leveraging basic In-Context Learning (ICL) and few-shot examples, devises the first iteration of the plan based on the sub-queries. (4) The plan is communicated to the expert agents and kicks off the swarming process - each expert agent starts its search. (5) The harmonizer synthesizes the information gathered via Swarm Intelligence (SI) and communicates it back to the task planning agent. (6) This enables the task planning agent to revise and optimize its previous plan. (7) The revised plan is then executed by the harmonizer agent and results are communicated back to the user.}
  \label{fig:FISHNET}
  \Description{Full System Architecture of FISHNET.}
\end{figure*}

Agent-Based Modeling (ABM) in finance has traditionally focused on simulating the interactions amongst market participants or simulating economic scenarios \citep{nie2024surveylargelanguagemodels}. In the realm of financial insight generation, traditional methods of knowledge graph construction have proliferated the research arena \citep{jiang2023evolutionknowledgegraphssurvey}. In this context, ABM to generate financial intelligence from multiple knowledge bases is a relatively unexplored realm of research for AI in finance.  Moreover, swarming Large Language Models (LLMs) in finance is a novel arena that explores the  collaboration of multiple LLM agents to achieve complex tasks \citep{jiao2023swarmgptcombininglargelanguage}. In the beginning, artificial Swarm Intelligence (SI) as a study drew its motivation by observing  emergent behaviors in nature such as swarming bees or schooling fish \citep{rosenberg2023conversationalswarmintelligencepilot}. In nature, the formation of  closed-loop systems among groups of
independent agents engenders high-level intelligence which
exceeds the capacity of the individual agents \citep{Rosenberg_2016} (Figure~\ref{fig:teaser}).  

Recently, with the rapid development of LLMs, studies on developing customized tools for expert usage \citep{cai2024largelanguagemodelstool}, leveraging LLMs as a central orchestrating agent \citep{rasal2024navigatingcomplexityorchestratedproblem}, or capitalizing on an LLM's planning capabilities \citep{valmeekam2023planningabilitieslargelanguage} have proliferated the research arena. While these streams are highly promising, firstly, there is a dearth of research into swarming LLMs to gather unknown intelligence from large financial datasets. 
Secondly, there is also a lack of studies that combine swarming, planning, orchestrating, or sub-querying into an agentic system, especially for use-cases in finance. 

Therefore, we propose a novel methodology, FISHNET (Financial Intelligence from Sub-querying, Harmonizing, Neural-Conditioning, Expert swarming, and Task planning), that amalgamates these existing methodologies to generate financial insights, during inference, from  a myriad of highly complex financial documents. Instead of fine-tuning LLMs for financial insight generation which can be very costly or prohibitive due to closed-source models \citep{wu2023bloomberggptlargelanguagemodel}, FISHNET proposes an effective alternative approach that achieves highly promising results (61.8\% successful execution rate) benchmarked against baseline scenarios such as Retrieval Augmented Generation (RAG; 5.0\% routing accuracy, 45.56\% expected recovery rate) and Generative Routing (55.7\% routing accuracy). We conduct rigorous empirical experiments, analyzing 98,034 U.S. regulatory filings submitted to the the Securities and Exchange Commission's (SEC) electronic database, EDGAR (Electronic Data Gathering, Analysis, and Retrieval) \citep{edgar} and the IAPD (Investment Advisor Public Disclosure) website \citep{iapd}. 
Instead of the traditional workflow of humans randomly searching for data points within documents or fine-tuning an LLM on these filings, FISHNET leverages the best of all worlds in swarming, harmonizing, neural-conditioning, and task planning for financial documents. This enables FISHNET to generate highly sophisticated responses to any complex query that is imposed such as \texttt{''Calculate Bank A's wallet share for single-stock swaps for the latest quarter``} (Figure~\ref{fig:FISHNET}).

\textit{Contributions.} In summary, our contributions are as follows:
\begin{enumerate}
\item We propose a novel agentic system to generate highly sophisticated financial intelligence from complex databases, providing flexibility compared to fine-tuning large open-source models for financial analysis.
\item Through our rigorous ablations, we empirically prove the importance of each agent and component involved in FISHNET. Moreover, FISHNET achieves a 61.8\% success rate in discovering, routing, and generating a solution for multi-filing, multi-entity queries. 
\item FISHNET is moreover, a novel means of employing ABM for database management or intelligence generation, contrary to traditional research of using ABM for market simulation. 

\end{enumerate}

\section{Related Work}

\subsubsection{Swarm Intelligence}
Multiple studies have delved into the subject of Swarm Intelligence (SI). In particular, AFSA (Artificial Fish Swarm Algorithm) strives to mimic the behavior of fish via a stochastic method that searches for a myriad of solutions in a randomized procedure \citep{Pourpanah_2022}. AFSA emulates three main behaviors that schools of fish exhibit: the first behavior is preying, the second is swarming, where fish tend to avoid danger by assembling in a group, and the third is executing without explicit leadership, via a biological organ that detects infinitesimal pressure changes in neighboring environments \citep{rosenberg2024collectivesuperintelligenceamplifyinggroup}. Thus,  AFSA initializes {\itshape n} randomly distributed artificial fish across the search space, where {\itshape n} is the population size. Each fish searches for positions with higher fitness values - ultimately,  at the end of each iteration, the position with the best fitness
value is compared with the previously archived solution in AFSA's bulletin board. If the new position is better, the bulletin will be updated. This iteration persists until the stop criterion is met \citep{Pourpanah_2022}.  \citet{rosenberg2024collectivesuperintelligenceamplifyinggroup} dives deeper in his recent study on collective super intelligence enabled via LLM agents.  These agents serve as the {\itshape organ} that can participate in simultaneous actions and communicate with each other.  A key characteristic of SI is that there is no central entity that orchestrates task management.  

\subsubsection{LLMs and Harmonization/Orchestration}

\citet{pichlmeier2024expertrouterorchestratingefficient} introduces the highly modular architecture of Expert Router that orchestrates the performance of multiple experts - diving deep into the capabilities of an LLM to serve as the central harmonizer of experts. Similarly, \citet{mohammadshahi2024leerooorchestratorelevatingllms} investigates the abilities of an LLM-based orchestrator that can effectively pick the expert agents for optimized task execution with regards to the input query. It is important to highlight that orchestration diverges from Mixture of Experts (MoE) modeling as the latter requires the router to select experts from the MoE layer; the MoE layer can then be distributed through multiple GPUs via Model Parallelism or Expert Parallelism techniques \citep{jiang2024mixtralexperts}.  In contrast, orchestration does not imperatively demand that the experts be housed in a single layer. The efficient role of LLMs as an orchestrator will be leveraged in FISHNET.

\subsubsection{LLMs and Neural-Conditioning}
\citet{li2023largelanguagemodelsunderstand} has delved into the emotional intelligence capabilities of LLMs and how contextual prompts can serve as a stimuli to significantly increase the performance of generative tasks. In contrast, \citet{wang2024negativepromptleveragingpsychologylarge} investigates the effects of negative emotional stimuli that also enhances the performance of LLMs. On a slightly different note, in \citet{zeng2024flowmindautomaticworkflowgeneration}, FlowMind empirically proves how grounding the LLM with context also helps improve its performance.  These studies spawn the potential of neurally conditioning or psychological priming an LLM to act in certain way.  In FISHNET, we psychologically prime our agents via positive and negative examples to act within expectations. 

\subsubsection{LLMs and Expert Agents}
The usage of tools or expert agents is a heavily explored arena of current academic research on LLMs \citep{qin2023toolllmfacilitatinglargelanguage}. For example, \citet{qin2023toolllmfacilitatinglargelanguage} introduces ToolLLaMA which has amalgamated 16,464 Application Programming Interfaces (APIs) that are leveraged in the construction of a valid solution. ToolLLaMA has proven to be effective in not only successfully solving complex tasks but also generalizing to new APIs. \citet{zeng2024flowmindautomaticworkflowgeneration} has empirically proven the remarkable performance of LLMs leveraging domain-specific tools or APIs.  
\citet{liu2024toolnetconnectinglargelanguage} estimates the usage of graphs in which each node is a tool - this invention, {\itshape ToolNet}, is scalable to thousands of tools and performs well on a myriad of multi-hop reasoning tasks. \citet{schimanski2024faithfulrobustllmspecialists} investigates the deployment of LLM specialists for Evidence-Based Question-Answering. 
This highly prevalent area of utilizing experts or tools will be expanded on in FISHNET, as we construct expert agents for each type of document.

\subsubsection{LLMs and Task Management}
Amongst LLMs' emergent capabilities that are not directly related to their original task of word sequence completion, their ability to conduct task management or sequential decision making is one of the prevalent areas of study \citep{valmeekam2023planningabilitieslargelanguage}. \citet{huang2024understandingplanningllmagents} conducts a comprehensive study into current research that investigates LLMs' planning capabilities. This paper categorizes planning abilities into one of the following five areas - task decomposition, multi-plan selection, external planner-aided planning, reflection and refinement, and memory-augmented planning. In this context, \citet{pallagani2024prospectsincorporatinglargelanguage} studies the integration of LLMs into Automated Planning and Scheduling (APS), leveraging the highly nuanced language capabilities of LLMs in traditional APS scenarios.  

\subsubsection{LLMs and Finance}
Generative AI has been applied to various NLP tasks within the financial sector \citep{Bi_2024}, from predictive modeling to sentiment analysis \citep{araci2019finbertfinancialsentimentanalysis, zmigrod2024buddiebusinessdocumentdataset, watson2024thingbadquestionh4r, 10.1145/3383455.3422520, zmigrod2024buddiebusinessdocumentdataset, watson2024directedcriteriacitationrecommendation}. Recently, BloombergGPT was released, trained on a myriad of documents, encompassing news, filings, press releases, and social media from Bloomberg archives \citep{wu2023bloomberggptlargelanguagemodel}. While impressive in its asserted capabilities, the model is entirely proprietary and very costly to train. A number of LLMs have been fine-tuned on finance-related tasks~\cite{zhang2023instruct, yang2023fingpt, liu2023fingpt, lee2024surveylargelanguagemodels, 10.1145/3539618.3591875}; 
however, rarely do they leverage LLMs for coding. Therefore, the motivation to derive financial insights from large amounts of data, without having to undergo the costly process of fine-tuning, remains strong.  This serves as the backbone for our efforts to compose FISHNET. 

\subsubsection{LLMs and Coding}
LLMs, especially in their potential for code generation, have seen considerable exploration and advancement~\cite{nijkamp2022codegen}.  These studies have explored chain of thought through code, as demonstrated in~\cite{liang2023code} for robotic programs, web browsing~\cite{nakano2022webgpt}, or table question-answering ~\cite{Watson_2023}. Recently, LLMs have shown the ability to construct modular code for visual question answering based on abstractions of high-level APIs ~\cite{subramanian2023modular}. FISHNET will leverage LLMs for coding workflow structures in a multi-agent framework.

\begin{table*}
  \caption{Overview of Investment Filings and Expert Agents. We enumerate for each filing who is required to file, what fields of interest are included, and how often the form must be filed. All of our filings can be amended by the filer, therefore all records must be reconciled to replace deprecated information. We provide references to the legal regulation and database where these filings are recorded. }
  \label{tab:filingtypes}
  \begin{tabular}{lcccccccc}
    \toprule
    Filing & Who? & What? & Frequency? & Structured? & Completeness? & Amended? & Source & Regulation\\
    \midrule
    13F & Investment Managers & Holdings & Quarterly & XML & Partial & Yes & \citep{edgar} & \citep{sec1934} \\
    N-CSR & '40 Act Funds & Annual Report &   Annual & Text & Multi-Fund & Yes & \citep{edgar} & \citep{ica1940}   \\
    N-CEN &  '40 Act Funds & Census & Annual & XML & Multi-Fund & Yes & \citep{edgar} & \citep{ica1940} \\
    N-PORT & '40 Act Funds & Holdings & Quarterly & XML & Single-Fund & Yes & \citep{edgar} &  \citep{ica1940} \\
    N-MFP & Money Market Funds & Holdings & Monthly & XML & Single-Fund & Yes & \citep{edgar} & \citep{ica1940} \\
    ADV & Investment Advisors & Entity Information & Annual & Text & Yes & Yes & \citep{iapd} & \citep{iaa1940}\\
    \bottomrule
  \end{tabular}
\end{table*}

\section{Methodology}

\subsection{Sub-querying Agent}
As a query is imposed to FISHNET, the sub-querying process consists of firstly assessing the query's quality or probability of hallucination, leveraging HalluciBot \citep{watson2024hallucibot, watson2024thingbadquestionh4r}. We utilize HalluciBot's binary classifications of hallucinatory or non-hallucinatory queries to continuously iterate towards a positive class transition. This process is essential since vague or poorly-worded queries, especially in searching across multiple databases, can significantly hinder the effectiveness of the downstream process \citep{jagerman2023queryexpansionpromptinglarge}. Then, the sub-querying process takes a query {\itshape Q} and outputs a set of {\itshape q} sub-queries: 
$\mathcal{Q}=\left\{q_0, q_1, \ldots, q_n\right\}$. To accomplish this, we leverage in-context learning (ICL) with an LLM to generate a sub-query {\itshape q} that is optimized for the task management or planning stage. 

\subsection{Task Planning Agent}
The Task Planning Agent's primary objective is to initialize the planning stage for each sub-query, collaborate with the Harmonizer Agent (Section~\ref{sec:harmonizer}), and update FISHNET's long-term memory with an optimized plan by leveraging the Swarm Intelligence (SI) generated by Expert Agents (Section~\ref{sec:swarms}).
Firstly, the Task Planning Agent takes the sub-queries {\itshape q} and starts devising a plan that will accomplish the resolution of each sub-query. The first iteration of this planning process is done solely by the Task Planning Agent, that has been given few-shot examples and In-Context Learning (ICL) to have a basic understanding of what each Expert Agent can accomplish. Secondly, the Task Planning Agent communicates with the Harmonizer Agent which synthesizes the SI compiled by multiple Expert Agents. The SI can encompass heretofore unknown information for the Task Planning Agent - as the latter has only been trained with few-shot learning whilst the Expert Agents are comprehensive and detailed experts for their respective filings. Therefore, the Task Planning Agent can revise its initial plan and devise an optimized plan, that will also be stored in long-term memory for future reference.

\subsection{Harmonizer Agent} \label{sec:harmonizer}
The Harmonizer Agent's prime focus is to communicate with each Expert Agent, synthesize the SI, and communicate it back to the Task Planning Agent; therefore, the initial plan can be refined using SI. The Harmonizer's next key role is to execute the optimized plan.  In other words, while the Expert Agents retrieve the relevant data from their respective databases, the Harmonizer can aggregate, multiply, divide, or perform any arithmetic actions, by following the optimized plan.

\subsection{Expert Agents} \label{sec:swarms}
We assign an Expert Agent for each type of U.S. regulatory filing in the document ingestion pipeline.  Each Expert will be fully capable of understanding all data fields in the filing, requirements, format, frequency of submission, and any other minute details. As regulatory filings differ immensely in all these aspects, a specialized agent for each filing will enable a truly modularized and expert-driven approach to assembling insights. For all filings, amendments to original filings are submitted on an ad-hoc basis - therefore, reconciliation across different timelines is key. 

\subsubsection{\textbf{N-PORT Agent}}
The N-PORT Agent fully comprehends and manipulates every single data field, ontology, hierarchy, filing entity (whether it is a fund, trust, or asset manager), and filing frequency for Form N-PORT. Form N-PORT or the Monthly Portfolio Investments Report is a mandatory filing for all registered management investment companies ('40Acts) or an Exchange-Traded Fund(ETF) that is organized as a Unit-Investment Trust (UIT). It is critical to note that this filing is not required for Money-Market Funds (MMFs). While filed every month, Form N-PORT is released every quarter regarding each fund's portfolio and each investment holding within the portfolio as of the last business day, or calendar day, of the last month in the quarter \citep{edgar}. 

\subsubsection{\textbf{N-MFP Agent}}
The N-MFP Agent precisely understands all key requirements and data fields of Form N-MFP. Form N-MFP is the monthly public reporting form used by money market funds required by section 30(b) of the Act and rule 30b1-7 under the Act (17 CFR 270.30b1-7). Similarly to Form N-PORT, Form N-MFP must report information about the fund and its portfolio holdings as of the last business day or any subsequent
calendar day of the preceding month \citep{edgar}.  

\subsubsection{\textbf{ADV Agent}}
The ADV Agent is skilled at understanding the Investment Adviser Public Disclosure (IAPD) website and the two different forms that comprise Form ADV. 
Form ADV is filed annually by an Investment Advisor (IA) that has to register with the SEC or state authorities. It houses different types of information from N-PORT or N-MFP - such as business ownership, employees, clients, or disciplinary information pertaining \citep{iapd}.

\subsubsection{\textbf{N-CEN Agent}}
The N-CEN Agent is an expert of Form N-CEN, an annual filing required for all registered investment companies ('40Acts), other than face-amount certificate companies.  The Agent understands the frequency of submission as well as the precise data fields that include provision of financial support, principal underwriters, fund type, or investments in foreign corporations \citep{edgar}. A notable point is that N-CEN can be submitted on a trust-level, with the trust referring to a group of funds. 

\subsubsection{\textbf{N-CSR Agent}}
The N-CSR Agent is knowledgeable of Form N-CSR, the certified shareholder report for registered investment management companies. Form N-CSR carries highly valuable data as it houses audited financial statements, including each fund's Statement of Operations, Statement of Assets and Liabilities, or the Portfolio of Investments.  The Agent is highly knowledgeable of every line item within the financial statements. 

\subsubsection{\textbf{13F Agent}}
The 13F Agent retains expertise for Form 13F, required for investment managers pursuant to Section 13(f) of the Securities Exchange Act of 1934. Rule 13f-1(a) stipulates that investment managers which exercise discretion for accounts holding Section 13(f) securities with a a certain market value needs to file 13F \citep{edgar}. Our Agent is highly skilled at understanding not only the ontology of 13F but also its unique filing structure. 









\begin{table*}
  \caption{Overview of Question Types. Each question is linked to the required Agent, and the median/total number of data points required to correctly answer each sample question. Hard questions require either: a) extraction from highly unstructured reports (\texttt{Q3}), or b) multiple agents (\texttt{Q0, Q1, Q2}).  }
  \label{tab:datacreation}
  \begin{tabular}{lcccc}
    \toprule
    Question & Agent(s)  & Answer Type & Median & Total \\
    \midrule
    \multicolumn{5}{c}{ \textit{Easy Questions}} \\
    \texttt{Q0: Get the aggregate cash equity positions...} & 13F & \texttt{float} & 1,319 & 462,792  \\
    \texttt{Q1: Get the aggregate option positions...} & 13F  & \texttt{float} & 273 & 103,905  \\
    \texttt{Q2: Get the regulatory AUM...} & ADV  & \texttt{float} & 1 & 111  \\
    \texttt{Q3: Get all funds managed by investment advisor...} & NCEN & \texttt{list} & 67.5 & 10,123  \\
    \texttt{Q4: Get all prime brokers for...} & ADV  & \texttt{list} & 35 & 8,139  \\
    \texttt{Q5: Get the country-level AUM for manager...} & NPORT, NCEN & \texttt{dataframe} & 1,168 & 164,732  \\
    \texttt{Q6: Get the money market net assets per fund...} & NMFP, NCEN & \texttt{dataframe} & 118 & 12,672  \\
    
    \midrule
    
    \multicolumn{5}{c}{ \textit{Hard Questions}} \\
    \texttt{Q0: Get all holdings of instrument type for fund...} & NPORT & \texttt{dataframe} & 42.5 & 44,215  \\
    \texttt{Q1: Calculate the counterparty split for advisor...} & NPORT, NCEN  & \texttt{dataframe} & 56 & 17,061  \\
    \texttt{Q2: Identity custom baskets expiring for...} & NPORT & \texttt{dataframe} & 24 & 2,513  \\
    \texttt{Q3: Get the total assets from the annual report...} & NCSR  & \texttt{float} & 30 & 7,755  \\
    
    \bottomrule
  \end{tabular}
\end{table*}

\section{Datasets}

\subsection{Preliminaries}
For FISHNET's training dataset, we have ingested 98,034 U.S. regulatory filings for six different types of filings from EDGAR and IAPD. These six filing types differ immensely in terms of ontology, format, submission requirements, citations, and data hierarchies. Moreover, they serve as critical sources of information for retail investors, brokerage firms, asset managers, and corporate entities. 

\subsection{N-PORT Dataset}
We ingest more than a full year's submission of N-PORT filings, starting from the 1st quarter of 2023 to the 2nd quarter of 2024. We have ingested a grand total of 40,372 original filings and 1,076 amendments that covers all portfolio holdings for all '40Act funds. We scrape these reports at a maximum throughput of 10 reports (SEC site limits for a single host). Once the filings are downloaded, we index each filing according to its unique ID.

\subsection{N-MFP Dataset}
Similarly to N-PORT, we ingest six quarters of filings into our data pipeline for N-MFP. This amounts to a grand total of 3,574 original filings and amendments. We extract and process the data into tables via indexing and chronological ordering. 

\subsection{ADV Dataset}
We crawl the IAPD website to ingest Form ADV for the past year, counting 15,292 filings or more than 685,000 PDF pages in total. 

\subsection{N-CEN Dataset}
 We ingest 2,909 N-CEN filings for the past 6 quarters. N-CEN is an annual filing that can also be filed on a group level. Therefore, we reconcile the unique trust-level identifiers so every filing is indexed in a relational database.

\subsection{N-CSR Dataset}
2,756 N-CSR filings are ingested in our dataset for the past six quarters. Form N-CSR varies greatly from filer to filer and is a highly unstructured type of filing. The data ingestion pipeline for N-CSR is one of most robust, leveraging multiple Natural Language Processing (NLP) and computer vision techniques to process the data from these filings. 

\subsection{13F Dataset}
For the past six quarters, we ingest 25,544 13F Holdings Reports, 5,353 13F Notices, 1,122 13F Holdings Reports Amendments and 36 13F Notice Amendments.  This amounts to 32,055 unique filings and amendments. We create citational knowledge graphs to track the relationships between filers.  

\section{Metrics}
We evaluate our experiments across three dimensions:
\begin{enumerate}
\item \textit{Retrieval} - We use R-Precision across all queries to account for variable retrieval sizes.
\item \textit{Routing} - We use accuracy to assess the LLM's effectiveness to route queries to the correct agents/tables in a zero-shot setting.
\item \textit{Agentic} - We use the success rate of creating an accurate solution to judge the full system architecture.
\end{enumerate}

\subsection{R-Precision}
R-Precision is a metric used to evaluate the performance of an information retrieval system. It is defined as the precision at $R$, where $R$ is the number of relevant documents for a given query. 
\begin{equation*}
\text{R-Precision} = \frac{|\text{Relevant Docs.} \cap \text{Retrieved Docs. at $R$}|}{R}
\end{equation*}
In the context of embedding-based retrieval, let $R$ be the total number of relevant documents for a query. Then R-Precision is calculated as the proportion of relevant documents retrieved in the top $R$ positions of the ranked list of retrieved documents. Therefore, R-Precision is equivalent to both the precision at the $R$-th position $(P@R)$ and the recall at the $R$-th position.





\subsection{Accuracy of Pathway Routing for Agents}

In evaluating the performance of pathway routing for agents, it is essential to consider the conditional accuracy of correctly identifying the agent and subsequently selecting the appropriate subtable. This two-step accuracy measure ensures that the overall routing process is evaluated comprehensively.

\subsubsection{Definition}

The accuracy of pathway routing is evaluated in two steps:
\begin{enumerate}
    \item \textbf{Agent Identification Accuracy:}~The proportion of instances where the correct agent, A, is identified, as ~$Acc_{agent}=P(A)$~
    \item \textbf{Sub-table Identification Accuracy:}~Given the correct agent, A, is identified, the proportion of instances where the correct sub-table, S,  is selected, as ~$Acc_{subtable}= P(S~|~A)$~
\end{enumerate}
The overall accuracy $Acc_{routing}$ can be considered as the product of these two accuracies, reflecting the conditional nature of the routing process. 
\begin{equation*}
Acc_{routing}=P(A,~S)=P(A)~\cdot~P(S~|~A)
\end{equation*}



\begin{table}[t!]
\centering
\caption{
Ablation Studies on Embedding Architectures.  Metrics are R-Precision. We report the recovery rate per filing datapoints, aggregated on each of our questions. The best results for each model are \underline{underlined}. Filing 13F only had a single table, therefore the Agent and Table level R-Precision are equivalent.
}
\label{tab:embedding-results}
\begin{tabular}{@{} l rrr @{}}
\toprule
& & \multicolumn{2}{c}{\textbf{R-Precision} (\%, $\uparrow$)} \\
\cmidrule{3-4}
\textbf{Filing} &  \texttt{Count} &  \texttt{Agent Level} & \texttt{Table Level}  \\
\midrule
\texttt{13F}        &   $600$    &     \underline{$48.7$}  &      \underline{$48.7$}   \\
\texttt{ADV}        &   $600$    &     $68.5$  &      \underline{$86.0$}   \\
\texttt{NCEN}       &   $300$    &     $75.1$  &      \underline{$80.6$}   \\
\texttt{NPORT}      &   1,800   &     $35.0$  &      \underline{$44.1$}   \\
\texttt{NCSR}       &   $300$    &     $16.8$  &      \underline{$54.9$}   \\
\texttt{NMFP}       &   $588$    &     $51.0$  &      \underline{$60.8$}   \\
\midrule
\textbf{Overall} & 4,188   &     $45.6$  &      \underline{$52.2$}   \\

\bottomrule
\end{tabular}
\end{table}


\begin{table}[t]
\centering
\caption{
Ablation Studies on Routing Architecture. The best accuracy for each task, per split, is \underline{underlined}. (E) refers to solely embedding-based strategies, while (G) refers to generative strategies. Hard questions that 
require more than two tables are evaluated sequentially and scored with partial credit for identifying at least one table right.
}
\label{tab:routing-results}
\begin{tabular}{@{} l rrr @{}}
\toprule
& \multicolumn{3}{c}{\textbf{Accuracy} (\%, $\uparrow$)} \\
\cmidrule{2-4}
\textbf{Strategy or Model} &  \texttt{EASY} &  \texttt{HARD} &  \texttt{AVERAGE} \\
\midrule
\multicolumn{4}{c}{Task: \textit{Agent Identification}} \\
\texttt{text-embedding-ada-002}~(E)  &  7.4 &              6.5 &     7.1 \\
\texttt{gpt-3.5-turbo-16k-0613}~(G)  &  \underline{$82.7$} &              \underline{$24.9$} &     \underline{$61.7$} \\
\texttt{gpt-4-0613}~(G)         &     67.6 &              10.6 &  46.9 \\
\texttt{Swarm}~(G) & 66.7 & 24.6 & 51.5 \\

\midrule
\multicolumn{4}{c}{Task: \textit{Sub-table Identification}} \\
\texttt{text-embedding-ada-002}~(E)  &  67.4 &              \underline{$74.9$} &     70.1 \\
\texttt{gpt-3.5-turbo-16k-0613}~(G)  & \underline{$99.1$} &              74.8 &     \underline{$90.3$} \\
\texttt{gpt-4-0613}~(G)         &     97.3 &              73.4 &  88.6 \\
\texttt{Swarm}~(G) & 98.9 & 69.4 & 88.2 \\

\midrule
\multicolumn{4}{c}{Task: \textit{Overall routing}} \\
\texttt{text-embedding-ada-002}~(E)  &  5.0 &              4.9 &     5.0 \\
\texttt{gpt-3.5-turbo-16k-0613}~(G)  &  \underline{$82.0$} &              \underline{$18.6$} &     \underline{$55.7$} \\
\texttt{gpt-4-0613}~(G)         &     65.8 &              7.8 &  41.6 \\
\texttt{Swarm}~(G) & 66.0 & 17.1 & 45.4 \\

\bottomrule
\end{tabular}%
\end{table}

\section{Experiments}

\subsection{Embeddings}\label{subsec:embeddings}
We used \texttt{text-embedding-ada-002} on each individual line item per table, as well on each table's and agent's metadata descriptions. Null value fields are dropped from the JSON representation. Across all agents and tables, this yields \textbf{5,052,421} vector embeddings at a dimensionality of 1,536 (31 GB of memory at \texttt{float32} precision). Embeddings are indexed by FAISS \citep{douze2024faisslibrary, johnson2019billion} using a Flat Euclidean (L2) Index for exact k-Nearest Neighbors (KNN).

\subsection{Question Creation \& Augmentation}

We define 7 easy and 4 hard question templates and their canonical code solution to generate random samples. Each sample is curated such that a complete answer is guaranteed - the sampling source is from the total space of valid inputs that yields solutions. 
The original templates are useful for measuring the repeatability and reliability of our agent framework. To test the robustness of our system architecture, we inject noise into the templates by using \texttt{gpt-3.5-turbo} as a query re-writer to generate 2 variations. Each question is tagged with the appropriate route for the agent and table. Finally, each canonical solution is executed to produce a solution and the relevant records required for it.

\subsection{Ablations and Results}

\subsubsection{\textit{Embedding Based (RAG)}} We embedded 5 million datapoints into 3 variable scopes: \texttt{global}, \texttt{agent}, and \texttt{table} level search spaces to test the most reliable retrieval  for data discovery. R-Precision, our retrieval metric, demonstrates that hierarchical embedding data is preferable to a flat global structure (Table~\ref{tab:embedding-results}). The highest gain in R-Precision is drawn from NCSR, as a bespoke embedding space allows for a cleaner KNN retrieval on a single table (vs. 3 tables at the agent level).
\begin{enumerate}
    \item \texttt{Global} - each data row is indexed into one unified index. By inundating the search space with millions of distractor points, \texttt{global} measures the accretive value of segmented indexes by \texttt{agents} or \texttt{tables}. However, given that our agents and tables are subsets of the \texttt{global} embedding space, we do not directly ablate this split and hypothesize that, at best, it will perform equivalent to the table-level ablation.
    \item \texttt{Agent} - each data row is indexed according to its respective agent. This partitions the search space to a single filing, measuring the intra-discriminative power of embeddings to retrieve when isolated from other filing and agent types. 
    \item \texttt{Table} - each data row is indexed per sub-table. Therefore, this hyper-local search space experiments with how well any RAG system could retrieve data within a single table, an alternative to SQL querying. In cases of perfect routing, this ablation explores the discriminative power of embeddings to discover valid records from irrelevant ones.
\end{enumerate}

\begin{table}[!th]
\centering
\caption{Agentic Success Rates (higher the better) on our dataset. We provide splits according to each filing, difficulty, and in aggregate. Note that when reporting on a filing-level, we evaluate the success rate equally if a query requires two filings. Therefore the total count is higher than the total actual number of queries (4,200 vs. 3,300).
}
\label{tab:final_results}
\begin{tabular}{@{} l r rrr  @{}}
\toprule
{} & {} & \multicolumn{3}{c}{\textbf{Success Rate (\%, $\uparrow$)}} \\
\cmidrule{3-5}

\textbf{Split} & \texttt{Count} & \texttt{Templated} & \texttt{Variegated} & Both  \\

\midrule

\multicolumn{5}{c}{\textit{Filing - Level}} \\

\texttt{13F}  &  600 &      $71.0$ &  $70.3$  &   $70.5$ \\

\texttt{ADV}  & 600 &     $70.5$ &  $72.3$  &   $71.6$ \\

\texttt{NCEN}  & 1,200 &     $80.2$ &  $89.3$  &   $86.3$ \\

\texttt{NCSR}  & 300 &     $3.0$ &  $25.0$  &   $17.7$ \\

\texttt{NMFP}  &  300 &     $100.0$ &  $93.4$  &   $95.6$ \\

\texttt{NPORT}  & 1,200 &     $39.3$ &  $50.0$  &   $46.4$ \\

\midrule 
\multicolumn{5}{c}{\textit{Easy Questions}} \\
\texttt{Q0}     & 300              &     $78.0$ &  $67.5$ &              $71.0$  \\
\texttt{Q1}     & 300              &     $64.0$ &  $73.0$ &              $70.0$  \\
\texttt{Q2}     & 300      &   $85.0$ &   $77.5$ &      $80.0$ \\
\texttt{Q3}     & 300      &     $94.0$ &  $99.0$ &               $97.3$ \\
\texttt{Q4}     & 300      &     $56.0$ &  $67.0$ &               $63.3$ \\
\texttt{Q5}     & 300              &     $85.0$ &  $92.5$ &              $90.0$  \\
\texttt{Q6}     & 300              &     $100.0$ &  $93.4$ &              $95.6$  \\

\textit{All Easy} & 2,100 & $80.2$ & $81.4$ & $81.0$ \\

\midrule 
\multicolumn{5}{c}{\textit{Hard Questions}} \\

\texttt{Q0}      & 300    &     $9.0$ &  $21.5$ &               $17.3$ \\

\texttt{Q1}     & 300             &     $42.0$ &  $72.5$ &              $62.3$  \\
\texttt{Q2}     & 300             &     $21.0$ &  $13.5$ &              $16.0$  \\

\texttt{Q3}    & 300      &     $3.0$ &  $25.0$ &               $17.6$ \\

\textit{All Hard}  & 1,200                &     $18.8$ &  $33.1$ &              $28.3$  \\

\midrule
\textbf{Overall}  &  3,300 &   $57.8$ &  $63.8$ &               $61.8$ \\

\bottomrule
\end{tabular}
\end{table}

\subsubsection{\textit{Routing}} We explore routing based methods, both embedding and generative. The Harmonizer agent (Section~\ref{sec:harmonizer}) would route to the correct expert agent leveraging a FAISS vector database populated with the embeddings of descriptions (Section~\ref{subsec:embeddings}) for each agent. The Expert agent (Section~\ref{sec:swarms}) uses a database containing the embeddings of a list of tables and the schema affiliated with each agent to perform routing. They are implemented as LangChain\footnote{https://www.langchain.com} chained Retriever QA agents that treats the vector database as as retriever. To empirically assess the accuracy of pathway routing, we employ a confusion matrix to identify agents and conditional probabilities to select sub-tables. Table~\ref{tab:routing-results} has a breakdown between Agents, Tables, and the Overall routing accuracy. 

\begin{enumerate}
    \item \texttt{Embedding} - We route a query based on matching its embedding with that of the agents' personas or tables' descriptions in a RAG setting.
    \item \texttt{Generative} - We use \texttt{gpt-3.5-turbo-16k} and \texttt{gpt-4} to isolate and link a query to an agent or a table based exclusively on semantic generation.
    \item \texttt{Swarming} - Using the generative guided query as a starting point, we simulate 5 agents performing collective-decision making over 2-3 timesteps to converge on a final answer using both the semantic generation of the query, but also all agents' reasoning history.
\end{enumerate}

As expected, the easy questions score 82.7\% at correctly selecting the agent, 99.1\% correct in selecting the table, and 82.0\% correct in determining the overall route. Easy questions, as single-agent focused or simple joins, had a high accuracy when juxtaposed to hard questions. These more difficult questions require more complex understanding of deeply hierarchical data in NPORT, leading to high accuracy in table identification (74.9\%), yet low agent identification accuracy (24.9\%). The best performing model for pathway routing was \texttt{gpt-3.5-turbo-16k-0613}.

\subsubsection{\textit{Agentic}}
In Table~\ref{tab:final_results}, we see that FISHNET is able to achieve a success rate of agent, table, and data discovery of 61.8\%. However, easy questions score in aggregate 81.0\%, compared to hard questions at 28.3\%. Furthermore, we see that variations actually perform slightly better, as the noise injected by \texttt{gpt-3.5-turbo} allows for our experiments to have more flexibility in recovering the answer, rather than fail on the same subset of templates repeatedly. Furthermore, query rewriting allows for an LLM to clarify a question conditioned on the original question, creating better questions.

\section{Conclusion}
In conclusion, we propose FISHNET, a multi-agent system to generate sophisticated financial insights. FISHNET combines swarming, sub-querying, harmonizing, planning, and neural-conditioning in a single system, demonstrating robust performance of 61.8\% in crafting accurate solutions while navigating a complex, hierarchical  agent-table data structure. As a novel means of leveraging ABM for financial data analysis, FISHNET provides a tangible alternative to fine-tuning or database engineering. Future work on FISHNET can focus on fine-tuning for different types of datasets, possibly non-English financial datasets that differ greatly in format or semantics. 

\section*{Disclaimer}
This paper was prepared for informational purposes by the Artificial Intelligence Research group of JPMorgan Chase \& Co. and its affiliates ("JP Morgan'') and is not a product of the Research Department of JP Morgan. JP Morgan makes no representation and warranty whatsoever and disclaims all liability, for the completeness, accuracy or reliability of the information contained herein. This document is not intended as investment research or investment advice, or a recommendation, offer or solicitation for the purchase or sale of any security, financial instrument, financial product or service, or to be used in any way for evaluating the merits of participating in any transaction, and shall not constitute a solicitation under any jurisdiction or to any person, if such solicitation under such jurisdiction or to such person would be unlawful.







\bibliographystyle{ACM-Reference-Format}
\bibliography{sample-base}


\end{document}